\documentclass{bmvc2k}

\usepackage{multirow}
\usepackage{adjustbox}
\usepackage{color,soul}
\usepackage{float} 
\usepackage{array}
\usepackage{amsmath}
\usepackage{amsfonts}
\usepackage{soul}
\usepackage[utf8]{inputenc} 
\usepackage[T1]{fontenc}    
\usepackage{hyperref}       
\usepackage{url}            
\usepackage{booktabs}       
\usepackage{amsfonts}       
\usepackage{nicefrac}       
\usepackage{microtype}      
\usepackage{lipsum}
\usepackage{fancyhdr}       
\usepackage{graphicx}       
\usepackage{algorithm}
\usepackage{algpseudocode}
\usepackage{xcolor}
\usepackage{wrapfig}
\graphicspath{{media/}}     
\usepackage{enumitem}
\usepackage{parskip}
\usepackage{algpseudocode}
\usepackage{adjustbox}
\usepackage{varwidth}
\usepackage{multicol}
\usepackage{bbm}
\usepackage{tabularx}
\usepackage{caption}
\usepackage{subcaption}
\usepackage{indentfirst}
\newcolumntype{P}[1]{>{\centering\arraybackslash}p{#1}}
\newcolumntype{M}[1]{>{\centering\arraybackslash}m{#1}}

\makeatletter
\def\blfootnote{\xdef\@thefnmark{}\@footnotetext}
\makeatother

\title{LoRAX: LoRA eXpandable Networks for Continual Synthetic Image Attribution}

\addauthor{Danielle Sullivan-Pao}{danielle.sullivan@ll.mit.edu}{1}

\addauthor{Nicole Tian}
{nicole.tian@yale.edu}{2*}

\addauthor{Pooya Khorrami}
{pooya.khorrami@ll.mit.edu}{1}

\addinstitution{
 MIT Lincoln Laboratory\\
 Lexington MA, USA
}

\addinstitution{
 Yale University\\
 New Haven CT, USA
}

\runninghead{Sullivan-Pao, Tian, Khorrami}{LoRAX}

\begin{document}

\maketitle
\begin{abstract}
As generative AI image technologies become more widespread and advanced, there is a growing need for strong attribution models. These models are crucial for verifying the authenticity of images and identifying the architecture of their originating generative models—key to maintaining media integrity. However, attribution models struggle to generalize to unseen models, and traditional fine-tuning methods for updating these models have shown to be impractical in real-world settings. To address these challenges, we propose LoRA eXpandable Networks (LoRAX), a parameter-efficient class incremental algorithm that adapts to novel generative image models without the need for full retraining. Our approach trains an extremely parameter-efficient feature extractor per continual learning task via Low Rank Adaptation. Each task-specific feature extractor learns distinct features while only requiring a small fraction of the parameters present in the underlying feature extractor's backbone model. Our extensive experimentation shows LoRAX outperforms or remains competitive with state-of-the-art class incremental learning algorithms on the Continual Deepfake Detection benchmark across all training scenarios and memory settings, while requiring less than 3\% of the number of trainable parameters per feature extractor compared to the full-rank implementation.\footnote{\url{https://github.com/mit-ll/lorax_cil}}
\end{abstract}

\blfootnote{* - Work done when author was an intern at MIT Lincoln Laboratory. DISTRIBUTION STATEMENT A. Approved for public release. Distribution is unlimited. This material is based upon work supported by the Department of the Air Force under Air Force Contract No. FA8702-15-D-0001. Any opinions, findings, conclusions or recommendations expressed in this material are those of the author(s) and do not necessarily reflect the views of the Department of the Air Force. © 2024 Massachusetts Institute of Technology. Delivered to the U.S. Government with Unlimited Rights, as defined in DFARS Part 252.227-7013 or 7014 (Feb 2014). Notwithstanding any copyright notice, U.S. Government rights in this work are defined by DFARS 252.227-7013 or DFARS 252.227-7014 as detailed above. Use of this work other than as specifically authorized by the U.S. Government may violate any copyrights that exist in this work.}

\section{Introduction}
\label{sec:intro}

The rapid democratization and advancement of image generation technologies emphasize the need for attribution models to verify the authenticity of visual content.  These models are vital for determining the origin of images, classifying their authenticity or identifying their generating model's architecture. Furthermore, by predicting the generative model behind a synthetic image, media forensic analysts can effectively investigate incidents and uncover coordinated disinformation campaigns. This capability is essential for upholding media integrity, national security, institutional trust, and public confidence in media sources.

One of the primary challenges of developing attribution models in the synthetic image (deepfake) generation/classification cat-and-mouse game is the fast-paced development of novel generation models \cite{dfd_survey}. These frequent advancements expand the set of possible source architectures; necessitating ongoing retraining or fine-tuning of attribution models \cite{icarl_dfd}. One approach, full retraining, is computationally demanding because it involves using the entire dataset for each class encountered, which may be impractical due to storage constraints and privacy issues \cite{privacy1, privacy2}. The second approach, iterative fine-tuning, suffers from catastrophic forgetting (CF) \cite{ewc}, a significant performance degradation on previously learned classes after training on new classes, unless the fine-tuning process incorporates mitigation techniques. Given that outputs from generative image models contain unique patterns, or "fingerprints", which can be utilized to infer details about the source model \cite{gan_fingerprints, fb_fingerprints}, we can formulate deepfake detection as a continual learning problem. This episodic style of fine-tuning classification models, where novel classes appear with time, is a well-studied problem known as class incremental learning (CIL) \cite{cil_survey}.


CIL algorithms aim to achieve the optimal balance in the stability-plasticity trade-off \cite{stab_plac}. Stability enables a model to retain knowledge from previous training episodes, while plasticity allows it to learn additional information in subsequent episodes. Recently, model-centric CIL algorithms, algorithms that expand the network backbone with each episode, have achieved state-of-the-art performance on continual learning benchmarks \cite{cil_survey}. However, when applying  existing models to the deepfake attribution problem, we have found that these models either suffer from catastrophic forgetting or result in a practically prohibitive explosion of model parameters. To address both of these issues, we leverage Low Rank Adaptation (LoRA) \cite{lora} and apply it to the problem of continual synthetic image attribution. Our proposed approach, LoRA Expandable Networks (LoRAX), trains a parameter-efficient feature extractor per CIL task. Each task's specific feature extractor is formed by applying the task-specific LoRA weight update to a single frozen backbone network. The task-specific feature extractors capture unique patterns left by each generative model, and the features from each task's extractor are ultimately fed to a unified classification head for attribution. LoRAX effectively avoids catastrophic forgetting by freezing both the underlying backbone model throughout training and the task-specific feature extractors at the conclusion of their respective training episodes.

In this paper, we present the following contributions: \vspace{-0.5em}
\begin{itemize}[noitemsep]

    \item We adapt the dynamic network CIL algorithms, DER and MEMO, to ConViT backbones. Our experimental results show that for each algorithm, ConViT backbones consistently outperform ResNet backbones with equal or fewer parameters, highlighting the importance of backbone selection in continual learning performance.
    
    \item We introduce LoRAX, a novel parameter-efficient class incremental learning algorithm.

    \item We complete an extensive set of experiments on the Continual Deepfake Detection (CDDB) benchmark to demonstrate the effectiveness of our LoRAX method across memory settings and CIL task datastreams. LoRAX is competitive with or outperforms other CIL algorithms across all tested learning scenarios and memory budgets.
\end{itemize}

\section{Related Work}
\label{sec:related_work}

Data-driven deepfake detection methods excel at identifying images generated by models included in their training set \cite{dfd_survey,ma_dfd}, but struggle to identify images generated by unseen techniques \cite{towards_gen}. This static, non-robust model training setup is impractical in the rapidly evolving field of generative AI \cite{gan_original, pix2pix, stylegan2, dall-e, vae, clip} and poses significant challenges for maintaining real-world classification accuracy. To address these challenges, researchers have successfully applied continual learning algorithms to the deepfake detection problem space \cite{icarl_dfd, cddb}, highlighting the potential of continual learning to enhance the robustness and adaptability of deepfake attribution models. Despite these advancements, there is still room for improvement in classification accuracy. In particular, adapting to novel generative techniques while mitigating forgetting requires further research and refinement.

\subsection{Class Incremental Learning}
Class Incremental Learning (CIL) algorithms handle continuously evolving data streams where new classes are introduced over time \cite{cil_survey}. The CIL data stream consists of task specific datasets, denoted as \( S = \{D^1, D^2, \ldots, D^N \} \). Each task dataset \( D^i \) represents a subset of data available at a specific time and is defined as \( D^i = \{ (x_i, y_i) \mid x_i \in X^i, y_i \in Y^i\} \), where \( X_i \) is the set of training instances from episode \( i \) and \( Y^i \) is the set of classes exclusively available in episode $i$. Importantly, the set of classes in each episode is non-overlapping \( Y^1 \cap Y^2 \cap \ldots \cap Y^N = \emptyset \), and the complete set of classes at the end of training is \( Y = \bigcup_{i=1}^N Y^i \).

In CIL, the model is updated episodically with each new task to incorporate additional classes.  Initially, model  \( F^1 \)  is trained to classify the original set of classes appearing in the first episode ( \( F^1: X^1 \rightarrow Y^1 \)). With each subsequent episode, the model is incrementally updated to include the new classes, evolving to  \( F^i: \bigcup_{t=1}^i X^t \rightarrow \bigcup_{t=1}^i Y^t \). This updating process is typically done with no or limited access to data from previous episodes. To help the model "remember" previously learned tasks, some CIL approaches use a subset of previous training data, known as exemplars, during fine-tuning. These exemplars are incorporated into the training data of future episodes. The goal of CIL is to continually adapt a single model to classify newly encountered classes while maintaining accuracy on previously learned classes and minimizing access to past data.

\subsubsection{Baseline Dynamic Network CIL Algorithms}
Recent research has witnessed a surge in the development of CIL algorithms \cite{ewc, icarl, der, memo, l2p, dytox, aws_lora} . Among these, model-centric \cite{cil_survey}  backbone expansion-based methods \cite{der, memo, dytox} have recently achieved state-of-the-art results. These approaches expand the backbone model to accommodate learning additional classes while minimizing interference with previously learned classes. Backbone expansion-based methods are particularly suitable for our specialized application of deepfake classification, as they do not rely heavily on pretrained networks, which may not generalize well to application-specific tasks \cite{cil_survey}. By dynamically expanding the network's architecture, backbone expansion-based methods ensure more robust performance in evolving data environments.
\vspace{-0.3cm}
\paragraph{Fine-tuning}
The fine-tuning approach, our baseline CIL algorithm, trains a single backbone model across all CIL tasks. With each task, it modifies the model by expanding the final classification head to include the latest task's new classes. With each CIL episode, the model weights from the previous task are retrained on the current task. No steps are taken to mitigate forgetting.
\vspace{-0.3cm}
\paragraph{DER}
Dynamically Expandable Representations (DER) is an early successful backbone expansion-based CIL algorithm \cite{der}. DER adds a backbone feature extractor for each CIL task. All of the extracted features are concatenated to form a “super feature”, which is then fed to a unified classifier. In addition to a traditional cross-entropy loss, DER incorporates an auxiliary loss. The auxiliary loss  uses the most recent task’s feature extractor to train a separate classifier to differentiate between all classes  in the current task \( Y^i \), and an additional class representing all previously seen classes \( \bigcup_{t=1}^{i-1} Y^t \). The goal of the auxiliary loss is to encourage the model to learn a diverse set of features from the existing set of feature extractors.\footnote{ DER uses the HAT \cite{hat} method to prune model parameters. HAT is hyperparameter sensitive and requires advance knowledge of the number of tasks, making it impractical in practice \cite{dytox}. It is not implemented in the paper's codebase, so we do not include it in our experiments. We denote our implementation of DER without the HAT method as "DER w/o P".}
\vspace{-0.3cm}
\paragraph{MEMO}
Based on the observation that shallow  neural network layers tend to extract general features, the Memory-Efficient Expandable Model (MEMO) \cite{memo} algorithm incorporates specialized blocks into a shared base for each CIL task. The specialized blocks efficiently integrate new tasks while leveraging the general features extracted by the shallow layers. MEMO also makes use of DER's auxiliary loss. 
\vspace{-0.3cm}
\paragraph{DyTox}Dynamic Token Expansion (DyTox) \cite{dytox} utilizes a transformer-based architecture tailored for CIL tasks. The algorithm features shared encoder and decoder layers, along with a set of dynamically expanding task-specific tokens. Each task token is prepended to the shared encoder's output features and fed into the shared decoder to produce a task-specific embedding. The set of task-specific embeddings is then used for classification. 
\vspace{-0.3cm}
\paragraph{Role of Exemplars}Each CIL algorithm we evaluate incorporates exemplars from past training episodes into the training dataset of subsequent tasks to help retain learned information. One commonly used exemplar selection strategy is herding \cite{icarl}. Herding selects "the most representative samples" by choosing those with features closest to their class's feature mean. 


\section{LoRA eXpandable Network}
Extending the work of existing dynamic network CIL algorithms, we define a CIL  algorithm that uses task-specific backbones to learn robust representations for each task while minimizing inter-task interference. We minimize the number of added parameters associated with each task's feature extractor by leveraging the parameter-efficient fine-tuning technique Low Rank Adaptation (LoRA)\cite{lora}.

\begin{figure}[t]
    \centering
    \includegraphics[width=\textwidth, trim={0 3cm 0 1.5cm}, clip]{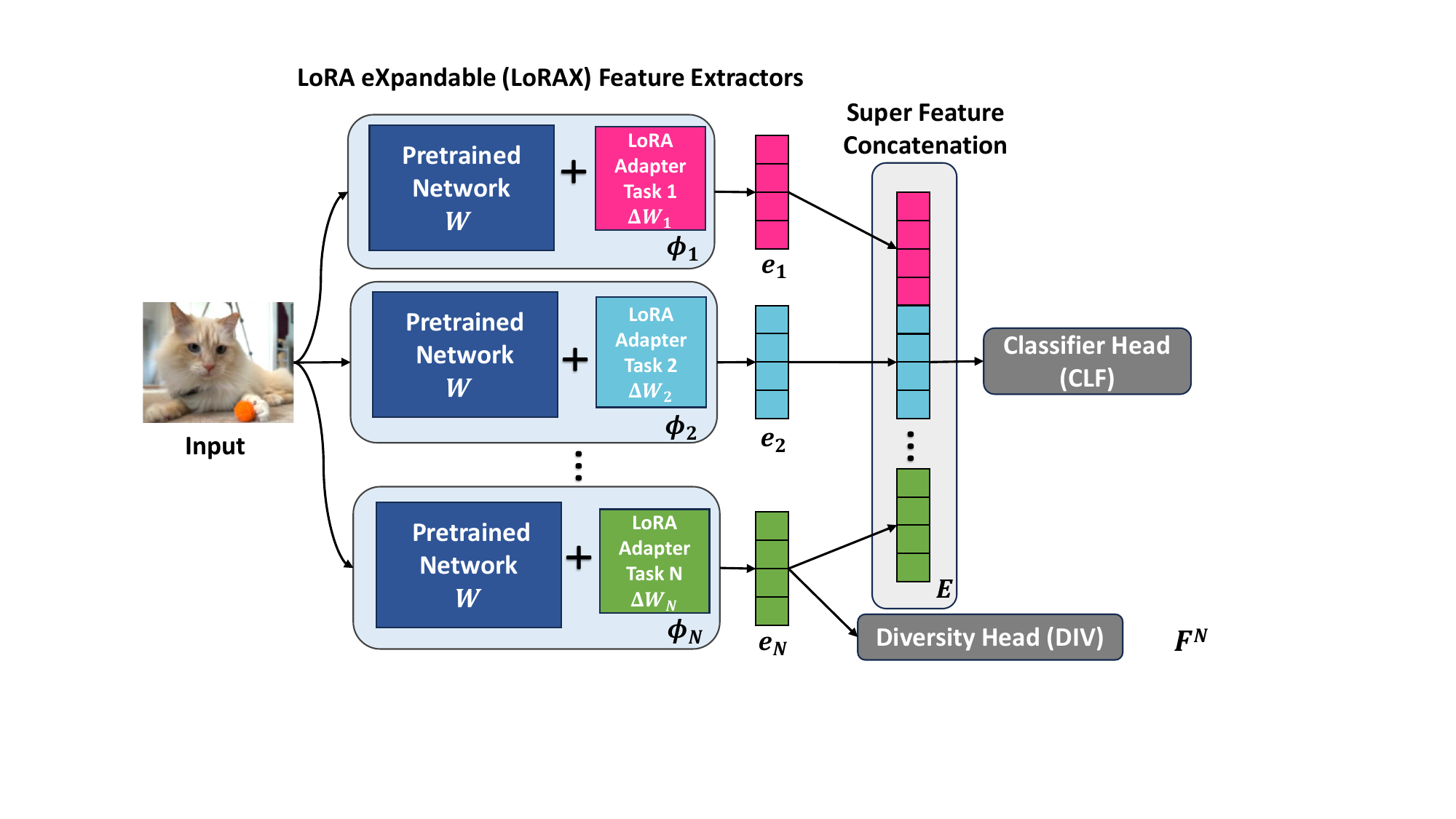} 
    \newline
    \caption{\textbf{LoRA eXpandable Network:} For each task $i$, a LoRA adapter $\Delta W_i$ is applied to the pretrained backbone network $W$ to form feature extractor $\phi_i$. Input images are passed through each feature extractor, and their output embeddings are concatenated all to form superfeature $E$. $E$ is passed to classifier head $CLF$ to predict image attribution. During training, the most recent task's feature $e_i$ is extracted and passed to diversity head $DIV$ to reduce redundancy between new and old features. $F^N$ refers to the model after $N$ tasks.}
    \label{fig:image-b}
\end{figure}

\subsection{Parameter-Efficient Fine-tuning}
Fine-tuning updates a pretrained network for a specific downstream application; requiring the storage of weight updates for every parameter in the network.  Inspired by the hypothesis that model weight updates have a low “intrinsic rank” \cite{intrinsic}, the LoRA algorithm \cite{lora} is defined such that for a pretrained weight matrix $W \in \mathbb{R}^{d \times k}$ with weight updates $\Delta W$, $\Delta W$ can be represented by the low-rank decomposition $\Delta W = BA$, where $B \in \mathbb{R}^{d \times r}$, $A \in \mathbb{R}^{r \times k}$, and the rank $r \ll \min(d,k)$. This  low rank representation greatly reduces the number of parameters needed to store the weight updates to the original network $W$ when $r \ll k$. 
\begin{equation}
\Phi_i = W + \Delta W_i  = W + BA
\label{eq:lora}
\end{equation}
LoRA fine-tuning initializes low-rank matrices $A$ and $B$ for each specified section of the network. During the model training, only the $A$ and $B$ matrices are updated while the underlying backbone model remains unchanged. The fine-tuned model is calculated by adding the product of the $A$ and $B$ matrices to the original model weights \eqref{eq:lora}. The CoLoR CIL algorithm \cite{aws_lora} also utilizes LoRA to train a separate classifier for each CIL task. However, it relies on a heavily pretrained backbone model to select the task classifier for each input. As noted in Section \ref{sec:related_work}, dependencies on extensively pretrained networks often degrade performance in specific applications, leading us to exclude it from our analysis.

\subsection{LoRAX Algorithm}
The LoRAX ConViT model results use the suggested configuration with a LoRA rank of $r = 64$. Details on determining the selected LoRAX model are in the supplementary material.
\paragraph{Dynamic Feature Expansion}
The LoRAX CIL algorithm (shown in Fig. \ref{fig:image-b}) trains a feature extractor \( \Phi_i \) per CIL task to capture the unique fingerprints left by each generator. Each feature extractor network is trained by applying LoRA weight updates to pretrained model \(W\), \( \Phi_i = W + \Delta W_i\). We limit the number of model parameters associated with each feature extractor by only storing the LoRA weight update matrix,  \( \Delta W_i\) . To mitigate catastrophic forgetting, we freeze pretrained network \(W\)  throughout training and freeze each \( \Delta W_i\) at the conclusion of its respective CIL episode. Following DER’s feature expansion pattern, we concatenate the features extracted from each task’s feature extractor to generate the “super feature,” $E(x)$.
\begin{equation}
 E(x) = [\Phi_1(x), \Phi_2(x), \ldots, \Phi_N(x)] 
\label{eq:super_feat}
\end{equation}
$E(x)$ is fed to the unified classification head, CLF, for model attribution.
\begin{equation}
p_{\text{CLF}}(y|x) = \text{softmax}(\text{CLF}(E(x)))\\
\label{eq:probs}
\end{equation}
%
\subsection{LoRAX Loss}
Following DER and MEMO, LoRAX uses a simple two-term loss function: cross-entropy loss \(L_{\text{CLF}}\) and diversity loss \(L_{\text{DIV}}\).
\vspace{-2em}

\begin{subequations}
\begin{align}
L &= L_{\text{CLF}} + \lambda L_{\text{DIV}} \label{eq:loss_sum}\\
L_{\text{CLF}}(x,y) &= \sum_{k=0}^{Y} - \mathbbm{1}{(y=k)} \log p_{\text{CLF},k} \label{eq:loss_clf}\\
L_{\text{DIV}}(x,y) &= \sum_{k=0}^{|Y^t| + 1} - \mathbbm{1}{(y=k)} \log p_{\text{DIV},k} \label{eq:loss_div}
\end{align}
\end{subequations}

\paragraph{Cross-Entropy Loss} The cross-entropy loss (Equation \ref{eq:loss_clf}) helps the model learn to classify the novel tasks encountered in the current training episode and mitigates forgetting on previously learned tasks by including exemplars from the current task’s training dataset. We expand the model’s unified classifier head at the start of each CIL episode to incorporate the task's novel classes and feature extractor. From the second task on, weights are inherited from the previous episode's CLF classifier.

\noindent\textbf{Diversity Loss} The diversity loss classifier is incorporated to minimize redundancy among the features extracted by each task adapter. This classifier is only required during training and is dropped at the conclusion of training each task. The hyperparameter $\lambda$ determines the weight of the diversity loss (Equation \ref{eq:loss_div}), impacting the balance between adapter feature diversity and classification accuracy. We ran a hyperparameter sweep for $\lambda = \{0.01, 0.1, 1.0\}$, and selected $\lambda=0.1$ for our experiments. 

\noindent\textbf{Exemplars} The LoRaX algorithm incorporates exemplar samples from previous episodes into its training process to prevent forgetting previously learned classes. Exemplars are selected via iCarl's \cite{icarl} herding process.
\section{Experiments}

Continual Deepfake Detection Benchmark (CDDB) \cite{cddb} is a deepfake detection benchmark for evaluating synthetic image detection/classification models in a continual setting. It was created by aggregating images from twelve well-known synthetic image classification datasets. The CDDB benchmark defines 3 training scenarios for continual learning model evaluation: easy (7 tasks), hard (5 tasks) and long (12 tasks). Each scenario defines a task order for training a CL model. Each task's dataset includes a set of real images and a set of synthetic images generated by known and unknown generative models. For tasks where the generative model is known, the real images correspond to the synthetic source’s training data. 

\subsection{Multi-real Setting}  
Each task in the CDDB dataset contains a set of real and synthetic images; consequently, each task in our CIL process results in an additional authentic image class. We employ a multi-real classification scheme that does not penalize for confusion between authentic classes (e.g. an authentic image from task $i$ is classified as an authentic image from task $j; j \ne i$). We calculate our performance metrics in a multi-real setting, such that the classification of an authentic image as any authentic image type is considered correct.

\subsection{Implementation Details} We implemented each CIL model in PyTorch \cite{pytorch}. Benchmark model code was based on the following open source implementations: DER \cite{pycil}, MEMO \cite{pycil}, DyToX \cite{cddb, dytox}. Our work extends the DER and MEMO implementations to include ConViT backbones using the PyTorch Image Model \cite{timm} ConViT implementation. All tested backbone models were initialized with ImageNet \cite{imagenet} pretrained weights. The LoRA fine-tuning component of our LoRAX model uses the HuggingFace \cite{huggingface} library. All models were trained on a single GPU (Fine-tuning, DER, LoRAX: NVIDIA A5000, A6000, L40 or A1000 GPU; MEMO, DyTox: NVIDIA Volta V100). We tuned hyperparameters for each CIL, backbone model combination on the CDDB hard scenario with 500 exemplars on a validation set of 15\% of the training data, which we used for all other scenario and memory settings.

\subsection{Evaluation Metrics} To evaluate a CIL algorithm's classification accuracy across a sequence of tasks and its ability to maintain performance on previously learned tasks, we track three continual learning metrics: Average Accuracy (AA), Average Accuracy at the Final Task (AAF) and Backward Transferability (BWT). AA represents the mean of average classification accuracies at each episode. AAF represents the mean classification accuracy across all tasks at the conclusion of the final episode. BWT measures the impact of learning new tasks on the performance of previously learned tasks, with a negative BWT value indicating performance degradation. Less negative BWT values indicate less forgetting. These metrics are computed on $n\times n$ matrix $A$, where $n$ is the number of tasks and entry $A_{i, j}$ is the accuracy of task $i$ after training on task $j$.
\begin{equation}
\setlength{\arraycolsep}{10pt}
\begin{array}{ccc}
\displaystyle \text{AA} = \frac{1}{n} \sum\limits_{i = 1}^n \left(\frac{1}{i} \sum\limits_{j = 1}^i A_{j, i}\right) &
\displaystyle \text{AAF} = \frac{1}{n} \sum\limits_{i = 1}^n A_{i, n} & 
\displaystyle \text{BWT} = \frac{1}{n - 1} \sum\limits_{i = 1}^{n-1} (A_{i, n} - A_{i, i}) \\
\text{(a)} & \text{(b)} & \text{(c)}
\end{array}
\label{eq:all}
\end{equation}

\subsection{Impact of Backbone Network on CIL Performance} 
\begin{figure}[t]
    \centering
        \includegraphics[scale=0.24]{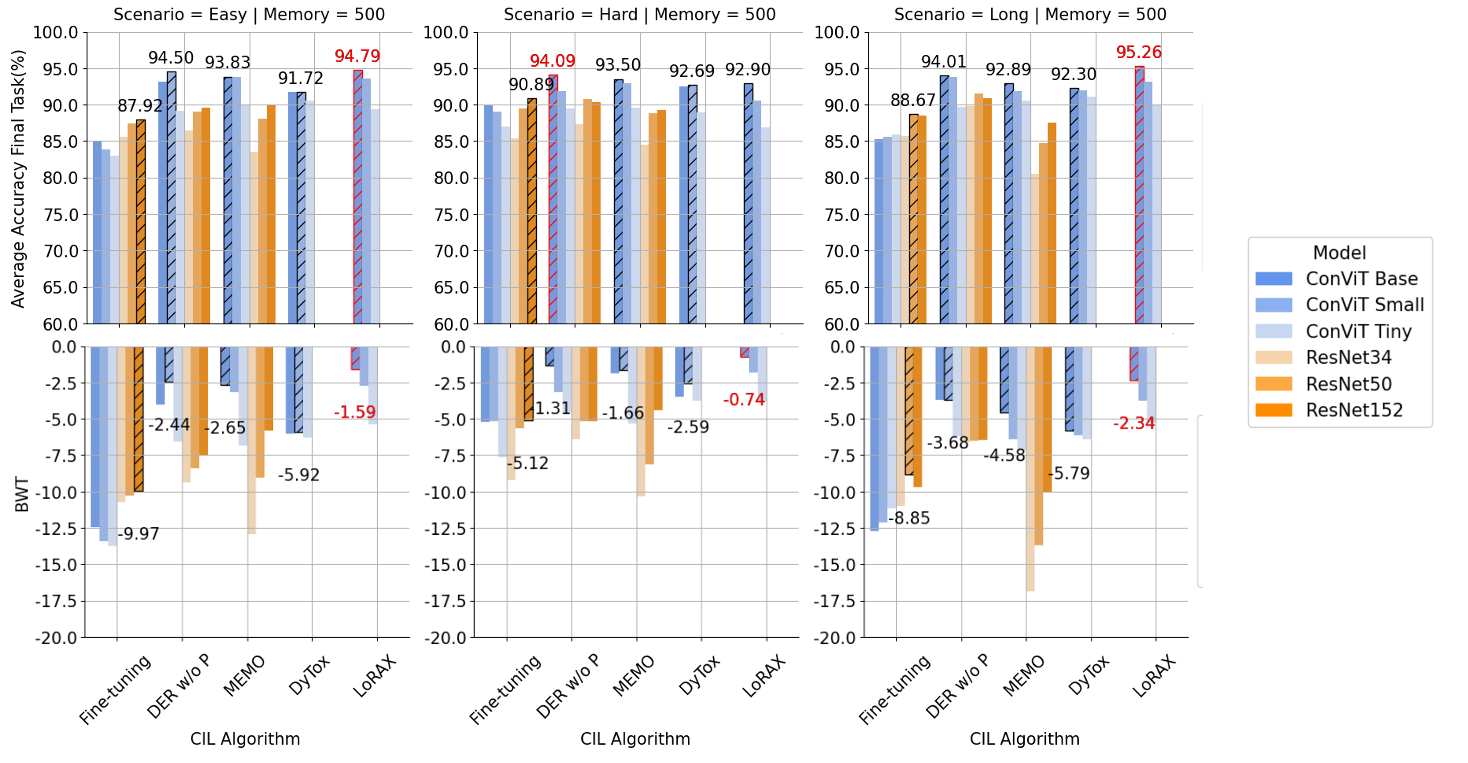}
        \newline
        \caption{\textbf{Comparison of Backbone Performance across CIL Algorithms.} Black hatching: top within CIL algorithm for the scenario \& memory setting; Red hatching: top overall for the scenario \& memory setting.}
        \label{fig:combined}
\end{figure}

We conducted a comparative analysis of the ResNet \cite{resnet} and ConViT \cite{convit} backbone architectures across the fine-tuning with exemplars, DER w/o P and MEMO CIL algorithms.\footnote{ DyTox was not included in this study due to the challenges of implementing its encoder, decoder, and task-specific tokens within a ResNet-based architecture in a straightforward manner. ResNet-based LoRAX was omitted due to poor performance.} Our results, as shown in Fig. \ref{fig:combined}, indicate that the choice of backbone model influences CIL algorithm performance. Within each tested dynamic network CIL algorithm, ConViT-based implementations consistently outperformed their ResNet-based counterparts of equivalent or lesser parameter counts in average accuracy, final average accuracy and backward transfer. Additionally, the ConViT Small model, 27.3 million parameters, outperformed the larger ResNet152 model, 58.1 million parameters, across all three metrics in both the DER w/o P and MEMO algorithms. This highlights its efficiency despite its smaller size.


\subsection{CDDB CIL Algorithm Evaluation}
Table \ref{table:cil_comparison} details the performance of every CIL method and backbone model combination across each CDDB scenario for the 500 exemplar memory setting. These results demonstrate that LoRAX is competitive across all evaluated CDDB scenarios, and outperforms all other tested CIL algorithms in each performance metric (AA, AAF, BWT) for both the easy and long scenarios. For the hard scenario, where LoRAX is not the leading algorithm, it performs within 1.2\% of the highest AAF score and 0.4\% of the highest AA score, while being the best performer for BWT.

\begin{figure}[t]
    \begin{center}
        \centering
        \includegraphics[scale=0.21]{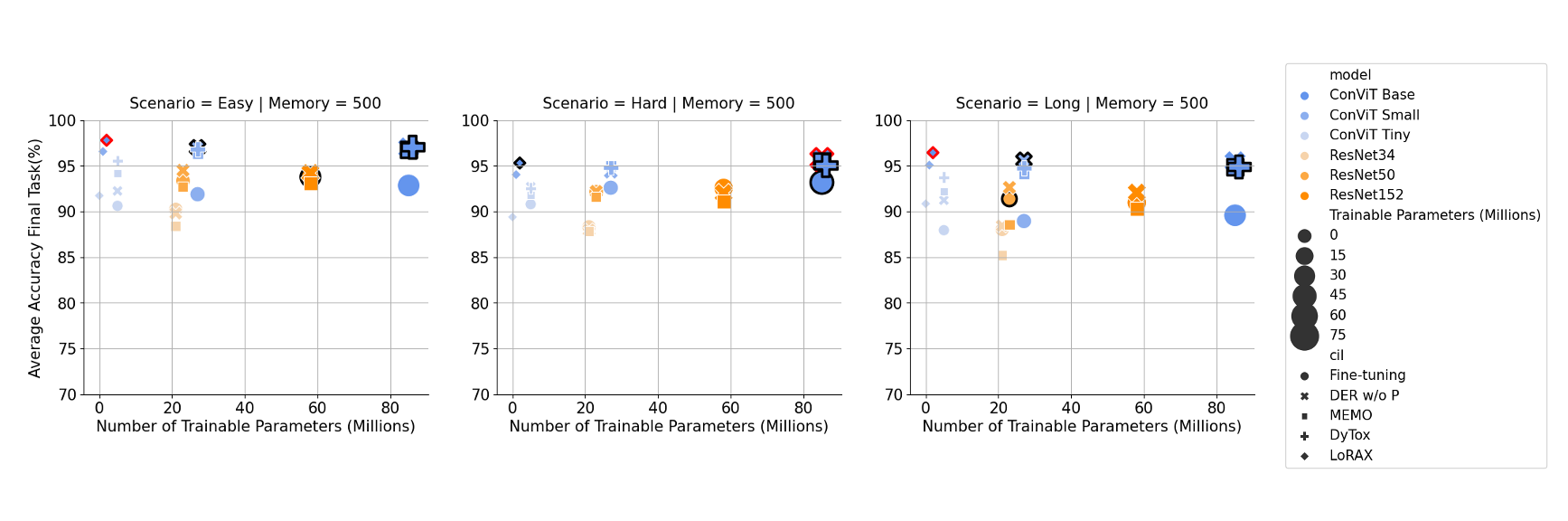}
        \caption{\textbf{CIL Algorithm Performance vs. Number of Trainable Parameters.} Black outline: top performer within CIL algorithm; Red outline: top overall.}
        \label{fig:scatterplot_500}
    \end{center}
\end{figure}

LoRAX maintains strong classification accuracy while requiring a relatively small number of trainable parameters per CIL episode. As seen in Fig. \ref{fig:scatterplot_500}, LoRAX requires the fewest number of trainable parameters across all tested CIL algorithms in each CDDB scenario. For the memory=500 setting, the LoRAX ConViT Base model excels, consistently ranking as a top-performing algorithm in terms of BWT, AA, and AAF, while requiring only a small fraction of the trainable parameters compared to other methods.\footnote{For a complete set of results across all CDDB memory settings shown in Figures \ref{fig:combined}, and \ref{fig:scatterplot_500}, refer to the supplementary material.}

\textbf{Oracle Comparison} To benchmark CIL algorithm performance against the joint training setting, we trained an Oracle model for each  backbone. Oracle models serve as an upper bound of classification model performance, where training data from all episodes is simultaneously available. At memory=500, the top-performing CIL algorithm in each CDDB scenario nearly matched the Oracle, falling short by just 1.5\% in both AA and AAF metrics. This robust performance highlights that even with limited rehearsal buffer, CIL is an effective approach for deepfake classification.

\textbf{LoRAX Improves Performance Over Fine-tuning with Exemplars} As seen in Table \ref{table:cil_comparison}, applying LoRA adapters to the fine-tuned ConViT backbone increases accuracy by 1-7\% AA and 1-10\% AAF across all scenarios, and this improvement holds true across all tested memory settings (see supplementary material). By concatenating feature dimensions from each task-specific feature extractor, LoRAX reduces forgetting across tasks relative to the baseline CIL framework fine-tuned with exemplars.

\begin{table}[th]
\begin{center}
\centering
    \begin{adjustbox}{width=0.85\textwidth}    
        \begin{tabular}{c|c|cccc|| cccc || cccc}
            \hline
            {} & {} & \multicolumn{12}{|c}{\textbf{500 EXEMPLARS}} \\
            \hline
            {} & {} & \multicolumn{4}{|c||}{Easy (35 exemplars/class at task N)} & \multicolumn{4}{|c||}{Hard (50 exemplars/class at task N)} & \multicolumn{4}{|c}{Long  (20 exemplars/class at task N)} \\
            \hline
             CIL & Model &    AA &    AAF &    BWT &  Params (M) &    AA &    AAF &    BWT &  Params (M) &    AA &    AAF &    BWT &  Params (M) \\
             \hline
            \multirow{6}{*}{Fine-tuning} & ConViT Base & 92.84 & 84.99 & -12.47 &   85.80 & \textbf{93.18} & 89.89 &  -5.20 &   85.80 & 89.58 & 85.23 & -12.71 &   85.80 \\
            & ConViT Small & 91.88 & 83.88 & -13.43 &   27.30 & 92.59 & 88.99 &  -5.18 &   27.30 & 88.95 & 85.59 & -12.13 &   27.30 \\
             & ConViT Tiny & 90.61 & 82.98 & -13.75 &    5.50 & 90.79 & 87.03 &  -7.63 &    5.50 & 87.95 & 85.90 & -11.15 &    5.50 \\
               & ResNet34 & 90.23 & 85.54 & -10.72 &   21.30 & 88.31 & 85.37 &  -9.23 &   21.30 & 88.03 & 85.66 & -11.00 &   21.30 \\
               & ResNet50 & 93.37 & 87.44 & -10.29 &   23.50 & 92.16 & 89.48 &  -5.66 &   23.50 & \textbf{91.38} & \textbf{88.67} &  \textbf{-8.85} &   23.50 \\
               & ResNet152 & \textbf{93.77} & \textbf{87.92} &  \textbf{-9.97} &   58.10 & 92.56 & \textbf{90.89} &  \textbf{-5.12} &   58.10 & 90.97 & 88.47 &  -9.71 &   58.10 \\
               \hline

            \multirow{6}{*}{DER w/o P} &
             ConViT Base & 96.94 & 93.09 &  -4.00 &   85.80 & \textbf{\textcolor{red}{\hl{95.69}}} & \textbf{\textcolor{red}{\hl{94.09}}} &  \textbf{-1.31} &   85.80 & 95.43 & \textbf{94.01} &  -3.69 &   85.80 \\
            & ConViT Small & \textbf{96.96} & \textbf{94.50} &  \textbf{-2.44} &   27.30 & 94.34 & 91.81 &  -3.15 &   27.30 & \textbf{95.60} & 93.82 &  \textbf{-3.6}8 &   27.30 \\
             & ConViT Tiny & 92.24 & 89.15 &  -6.57 &    5.50 & 92.70 & 89.49 &  -4.58 &    5.50 & 91.23 & 89.64 &  -6.17 &    5.50 \\
                & ResNet34 & 89.81 & 86.42 &  -9.36 &   21.30 & 87.98 & 87.34 &  -6.40 &   21.30 & 88.42 & 89.75 &  -6.61 &   21.30 \\
                & ResNet50 & 94.49 & 88.99 &  -8.40 &   23.50 & 92.19 & 90.79 &  -5.14 &   23.50 & 92.62 & 91.48 &  -6.53 &   23.50 \\
               & ResNet152 & 94.19 & 89.60 &  -7.46 &   58.10 & 92.01 & 90.33 &  -5.13 &   58.10 & 92.06 & 90.83 &  -6.45 &   58.10 \\
            \hline

            \multirow{6}{*}{MEMO} &
             ConViT Base & \textbf{96.79} & \textbf{93.83} &  \textbf{-2.65} &   85.80 & \textbf{95.54} & \textbf{93.50} &  -1.88 &   85.80 & \textbf{94.8}6 & \textbf{92.89} &  \textbf{-4.58} &   85.80 \\
            & ConViT Small & 96.36 & 93.79 &  -3.18 &   27.30 & 94.96 & 92.88 &  \textbf{-1.66} &   27.30 & 94.10 & 91.81 &  -6.39 &   27.30 \\
             & ConViT Tiny & 94.21 & 89.92 &  -6.82 &    5.50 & 91.82 & 89.58 &  -5.32 &    5.50 & 92.19 & 90.50 &  -7.17 &    5.50 \\
              &  ResNet34 & 88.38 & 83.53 & -12.93 &   21.30 & 87.84 & 84.54 & -10.35 &   21.30 & 85.22 & 80.52 & -16.89 &   21.30 \\
              &  ResNet50 & 92.70 & 88.05 &  -9.05 &   23.50 & 91.61 & 88.79 &  -8.13 &   23.50 & 88.55 & 84.70 & -13.68 &   23.50 \\
              & ResNet152 & 93.04 & 90.02 &  -5.79 &   58.10 & 91.10 & 89.26 &  -4.42 &   58.10 & 90.27 & 87.55 & -10.09 &   58.10 \\
              \hline

            \multirow{3}{*}{DyTox} &
             ConViT Base & \textbf{97.05} & 91.70 &  -6.03 &   64.50 & \textbf{95.06} & 92.45 &  -3.46 &   64.50 & \textbf{94.90} & \textbf{92.30} &  \textbf{-5.79} &   64.50 \\
            & ConViT Small & 96.83 & \textbf{91.72} &  \textbf{-5.92} &   20.90 & 94.79 & \textbf{92.69} &  \textbf{-2.59} &   20.90 & 94.75 & 91.94 &  -6.11 &   20.90 \\
            & ConViT Tiny & 95.55 & 90.51 &  -6.30 &  4.14 & 92.54 & 88.93 &  -3.75 &  4.14 & 93.74 & 91.12 &  -6.41 &  4.14 \\
            \hline

             \multirow{3}{*}{LoRAX} &
             ConViT Base & \textbf{\textcolor{red}{\hl{97.79}}} & \textbf{\textcolor{red}{\hl{94.79}}} &  \textbf{\textcolor{red}{\hl{-1.59}}} &    \textbf{\textcolor{red}{2.50}} & \textbf{95.30} & \textbf{92.90} &  \textbf{\textcolor{red}{\hl{-0.74}}} &    \textbf{\textcolor{red}{2.50}} & \textbf{\textcolor{red}{\hl{96.43}}} & \textbf{\textcolor{red}{\hl{95.26}}} &  \textbf{\textcolor{red}{\hl{-2.34}}} &    \textbf{\textcolor{red}{2.50}} \\
            & ConViT Small & 96.56 & 93.54 &  -2.72 &    \textbf{\textcolor{red}{1.40}} & 94.03 & 90.54 &  -1.80 &    \textbf{\textcolor{red}{1.40}} & 95.07 & 93.17 &  -3.76 &    \textbf{\textcolor{red}{1.40}} \\
            & ConViT Tiny & 91.73 & 89.39 &  -5.38 &    \textbf{\textcolor{red}{0.60}} & 89.39 & 86.85 &  -3.78 &    \textbf{\textcolor{red}{0.60}} & 90.84 & 89.88 &  -5.24 &    \textbf{\textcolor{red}{0.60}} \\
            \hline
        {} & {} & \multicolumn{12}{|c}{\textbf{ALL TRAINING DATA}}\\
        \hline
        \multirow{6}{*}{Oracle}& 
         ConViT Base & 96.52 & 93.99 & N/A &   85.80 & 93.92 & 91.47 & N/A &   85.80 & 95.71 & 94.28 & N/A &   85.80 \\
        & ConViT Small & 96.71 & 94.38 & N/A &   27.30 & 94.19 & 91.68 &  N/A &   27.30 & 95.99 & 94.36 & N/A &   27.30 \\
         & ConViT Tiny & 96.30 & 93.81 & N/A &    5.50 & 93.62 & 91.64 & N/A &    5.50 & 95.60 & 94.27 & N/A &    5.50 \\
          &  ResNet34 & 95.87 & 94.07 & N/A &   21.30 & 93.71 & 91.82 &  N/A &   21.30 & 95.14 & 94.11 & N/A &   21.30 \\
          &  ResNet50 & \textbf{\textcolor{blue}{98.11}} & \textbf{\textcolor{blue}{95.83}} & N/A &   23.50 & \textbf{\textcolor{blue}{95.69}} & 93.87 & N/A &   23.50 & 96.74 & 94.66 & N/A &   23.50 \\
          &  ResNet152 & 97.70 & 95.44 & N/A &   58.10 & 95.65 & \textbf{\textcolor{blue}{95.24}} &  N/A &   58.10 & \textbf{\textcolor{blue}{96.77}} & \textbf{\textcolor{blue}{95.34}} & N/A &   58.10 \\
        \hline
        \end{tabular}
    \end{adjustbox}
\end{center}
\vspace{0.5cm}
\caption{\textbf{CIL algorithm performance comparison for 500 exemplar memory setting.} \textbf{Top within CIL algorithm}, \textbf{\textcolor{red}{\hl{Top overall}}}, \textbf{\textcolor{red}{Fewest trainable params}},  \textbf{\textcolor{blue}{Top Oracle model}}.}
\label{table:cil_comparison}
\end{table}
\newpage

\vspace{-3cm}
\section{Conclusion}
In this paper, we propose LoRAX, a novel class incremental learning algorithm that leverages LoRA to train an extremely parameter-efficient feature extractor per class incremental learning task. Our task-specific feature extractors enable LoRAX trained models to recognize artifacts unique to each task while minimizing interclass learning interference. Moreover, by freezing each feature extractor following its respective CIL episode, LoRAX reduces catastrophic forgetting. Compared to the underlying ConViT Base backbone model, our suggested LoRAX model dramatically decreases the memory of each task-specific feature extractor. We evaluate our LoRAX method on the Continual DeepFake Detection Dataset and show it achieves competitive performance compared to a set of contemporary dynamic network CIL algorithms.


\section{Appendix}

\vspace{-0.5em}

\subsection{Symbols and Meanings}
\begin{table}[h]
\centering
\begin{tabular}{>{\raggedright\arraybackslash}p{3cm} p{10cm}}
\hline
\textbf{Symbol} & \textbf{Meaning} \\
\hline
$W$ & Pretrained backbone network \\
$\Delta W_i$ & Task $i$ LoRA Adapter \\
$F^i$ & Classifier network after training on task $i$ \\
$\Phi_i$ &  Task $i$ LoRA-based feature extractor, $W + \Delta W_i$ \\
$e_i$ & Embedding extracted from $\Phi_i$  \\
$E$ & Super feature  \\
$S$ & CIL datastream \\
$N$ & Total number of CIL episodes \\
$M$ & Exemplar buffer \\
$B$ & Exemplar budget (number of samples) \\
CLF & Classification head \\
DIV & Diversity head \\
\hline
\end{tabular}
\vspace{0.5cm}
\caption{Symbols and their meanings}
\label{table:symbols}
\end{table}

\vspace{-1.3em}

\subsection{LoRAX Algorithm}
\begin{algorithm}
\caption{LoRAX  Algorithm (Training)}
\begin{algorithmic}[1]
\State \textbf{Input:} Data stream $S$ composed of $N$ tasks, pre-trained network $W$
\State \textbf{Initialize:} Exemplar buffer $M \gets \{\}$,  Freeze $W$
\For{each task $i$ in $S$}
  \State Initialize $\Delta W_i $ \Comment{Initialize task $i$'s LoRA matrices}
   \State Expand CLF to accommodate $e_i$ and $Y^i$ \Comment{Expand classifier head }
    \If{ $i >$1}
        \State Initialize DIV \Comment{Initialize diversity head }
    \EndIf
   \State Train $F^i$ on $D^i \bigcup M$  \Comment{Train on task $i$ data and exemplars }
     \State Freeze $\Delta W_i $ , drop DIV
    \State Update $M$ with samples from task $i$
    \If{ $|M| >$ $B$}
        \State Use herding to remove samples such that $|M|\leq$  $B$ 
    \EndIf
\EndFor
\State \textbf{Output:} $F^N$
\end{algorithmic}
\end{algorithm}

\vspace{-0.75em}

\begin{algorithm}[H]
\centering
\begin{algorithmic}[1]
\State \textbf{Input:} Image $x$ 
\For{$i = 1$ to $N$} \Comment{Loop through each task extractor}
    \State $e_i = Wx + \Delta W_i x$ \Comment{Feature $i$}
\EndFor
\State $E(x)=[e_1 , e_2, ... , e_N]$ \Comment{Super feature}
\State $\hat{y}=\arg\max(\text{CLF}(E(x)))$ \Comment{Classification}
\State \textbf{Output:} $\hat{y}$
\end{algorithmic}
\caption{LoRAX Forward Pass (Inference) }
\label{alg:training}
\end{algorithm}

\subsection{The importance of high-frequency details}
\begin{figure}[H]
    \begin{center}
        \centering
        \includegraphics[scale=0.2]{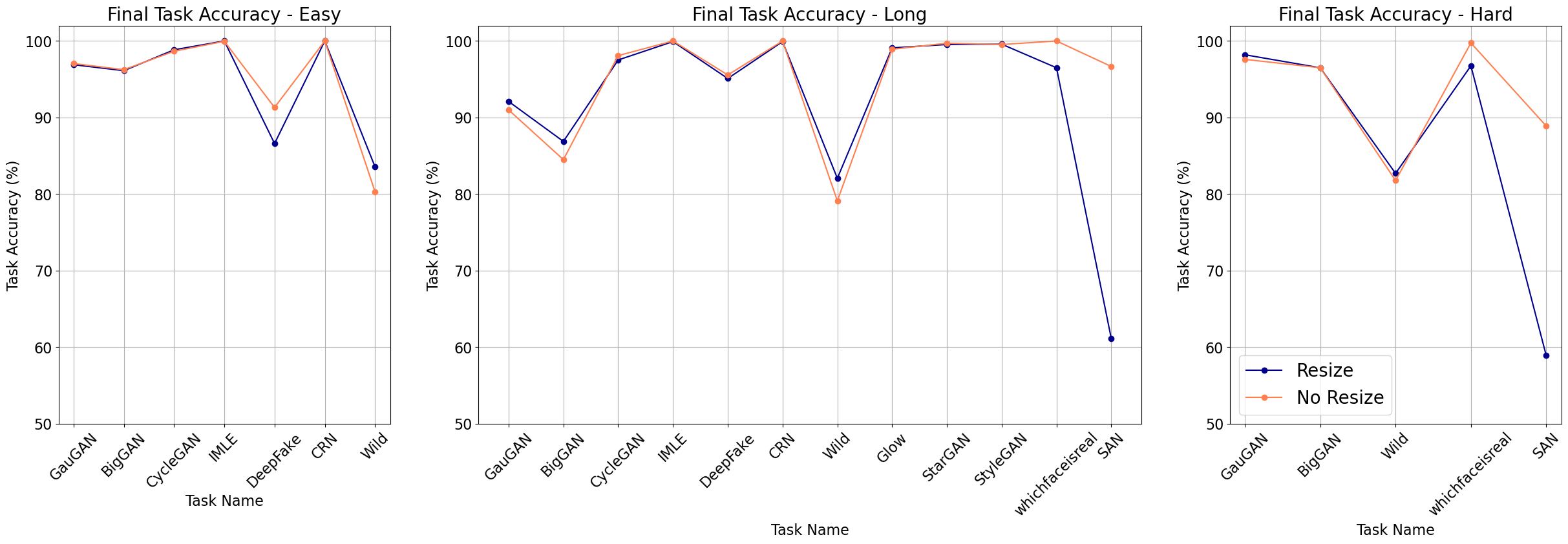} \newline
        \label{fig:resizing}
        \caption{\textbf{Effect of Image Resizing on Final Task Accuracy (ConViT Base, r=64, $M$=500)} The final accuracy of each task with (blue) and without resizing (coral) after training on the final CIL task. Tasks are plotted in the order they appear in the data stream (e.g. GauGAN is task 1 of the Easy scenario). }
    \end{center}
\end{figure}

Each of the CIL algorithms we evaluated  incorporate resizing and cropping in the preprocessing stages of their official open source implementations. These steps are designed to adjust an image's dimensions to conform to the expected model input size while attempting to preserve the semantic content of the images. However, our experiments demonstrate that the interpolation used in the resizing process eliminates high-frequency details that are crucial for identifying specific synthetic image classes. These findings are in agreement with previous research \cite{for_now, resize}. Notably, when resizing was included in preprocessing for the LoRAX model with an exemplar budget of $M=500$, there was a significant decline in classification performance on the SAN dataset—dropping by 35.56 and 30.00 percentage points in the long and hard scenarios. This decline can be attributed to the removal of high-frequency details present in SAN generated images \cite{san}, which were lost during resizing. To ensure a fair comparison across all tested CIL algorithms, we omit resizing from all preprocessing steps, utilizing only cropping to align images with a model's expected input sizes.

\subsection{Exemplar-free CDDB Evaluation}
To evaluate the performance of our exemplar-based CIL algorithms against an exemplar-free CIL algorithms, we tested Learning to Prompt (L2P), a leading exemplar-free CIL algorithm \cite{l2p}. L2P dynamically selects and prepends learnable prompts from a prompt pool to input image embeddings. The augmented embeddings are passed to a pre-trained encoder for classification. We ran the top performing pre-trained model from the L2P paper (ViT-B/16) on the CDDB easy, hard, and long scenarios. This model was pretrained on ImageNet and yields SOTA performance on CIFAR-100 \cite{cil_survey}. However, this model did not generalize well to the CDDB dataset as seen in Table \ref{table:l2p}.
\begin{table}[b]
\begin{center}
\centering
    \begin{tabular}{|ccc || ccc || ccc|}
        \hline
        \multicolumn{9}{|c|}{L2P Results} \\
        \hline
        \multicolumn{3}{|c||}{Easy} & \multicolumn{3}{c||}{Hard} & \multicolumn{3}{c|}{Long} \\
        \hline
        {AA} & {AAF} & {BWT} & {AA} & {AAF} & {BWT} & {AA} & {AAF} & {BWT} \\
        \hline
        76.79 & 75.16 & -4.96& 66.97 & 63.34 & 0.30 & 72.03 & 69.41 & -11.87\\
        \hline 
    \end{tabular}
\end{center}
\vspace{0.5cm}
\caption{Exemplar-free L2P (ViT-B/16) CDDB Results  }
\label{table:l2p}
\end{table}
\subsection{Selecting Top Performing LoRAX Model}
The LoRAX CIL algorithm is notably straightforward, and requires minimal hyperparameter tuning. To establish our recommended configuration for the LoRAX model, we conducted a comprehensive sweep of two key LoRA settings: the adapter configuration, which determines the layers where LoRA is applied, and the rank, which specifies the inner dimension of the LoRA update matrices.
\begin{figure}[t]
    \begin{center}
        \centering
        \includegraphics[scale=0.3]{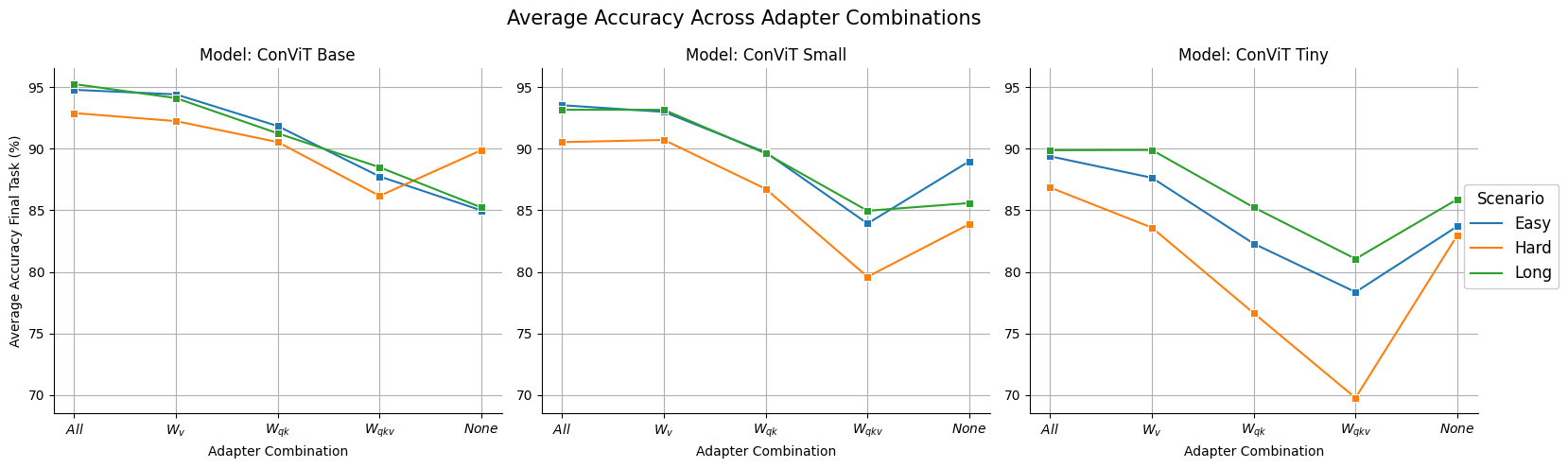} \newline
        \caption{ \textbf{LoRAX AAF Across Adapter Combinations (Memory = 500)} The weight matrices we applied adapters to is shown on the x-axis. All refers to $W_v, W_{qk}, W_{qkv}$, and the ConViT positional embeddings.}
        \label{fig:adapters}
    \end{center}
\end{figure}

\subsubsection{Optimal Adapter Combination}
We performed an adapter configuration sweep to identify the optimal transformer block matrices of the ConViT model for applying LoRA updates (see Figure \ref{fig:adapters}). We grouped blocks by following the experimentation outlined in the original LoRA paper \cite{lora}, resulting in the following adapter combinations: $W_v, W_{qk}, W_{qkv}$, and All ($W_v, W_{qk}, W_{qkv}$ and the matrices representing ConViT location information). We found that applying adapters to all the matrices in the attention modules consistently generated the highest average accuracy across all tested ConViT model and CDDB scenario combinations. While including more components of the backbone model in the LoRA configuration increases the number of trainable parameters, we determined that the performance trade-off justified this increase. Even with additional modules included in the LoRA configuration, LoRAX feature extractors are a small fraction of the size of their full-rank counterparts.

\subsubsection{Optimal Adapter Rank}
\begin{figure}[ht]
    \begin{center}
        \centering
        \includegraphics[scale=0.3]{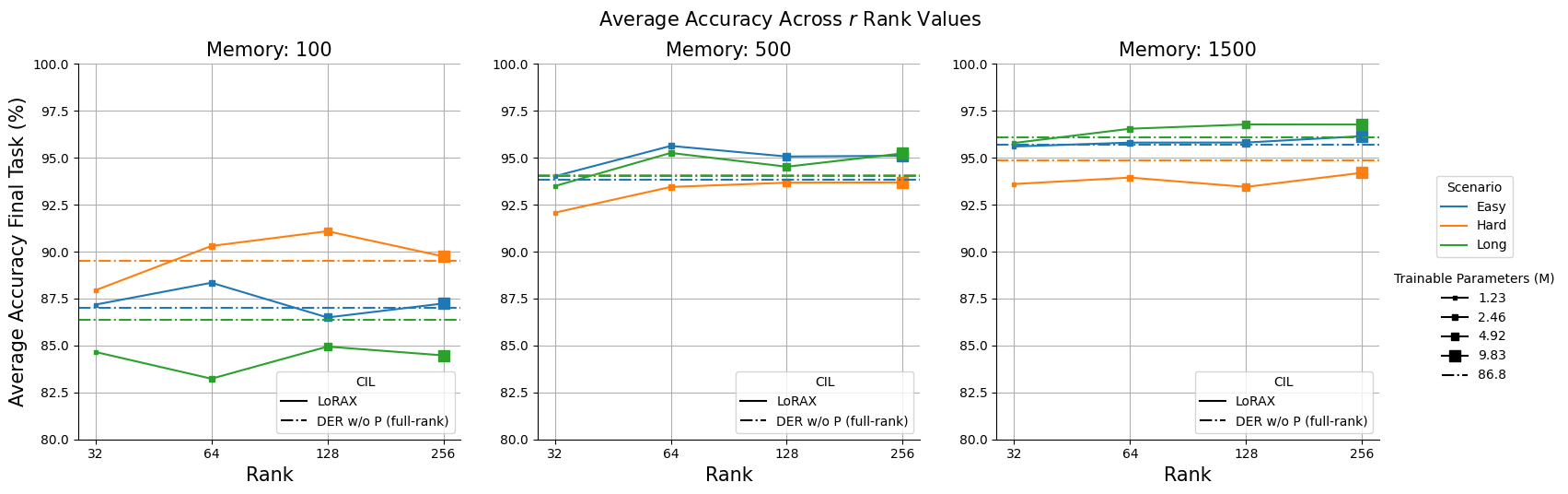}\newline
        \caption{\textbf{AAF across different rank values (Memory = 500)} All experiments were performed with the top-performing adapter combination in Figure \ref{fig:adapters}. Marker sizes for each ConViT LoRAX model are scaled based on number of trainable parameters. We compare against the DER w/o P benchmark plotted for each scenario and memory setting with the dotted line, since our feature concatenation method draws on the DER feature concatenation method. We use the DER w/o P benchmark as an ablation study of our CIL algorithm without adapters.}
        \label{fig:ranks}
    \end{center}
\end{figure}
To identify our recommended LoRAX rank value (r), we performed a hyperparameter sweep, doubling the rank from r=32 to r=256 at each step (see Figure \ref{fig:ranks}). We completed this sweep for each CDDB scenario and memory setting combination. In 8 out of 9 tested configurations,  the LoRAX model $r = 64$ outperformed  its  $r = 32$  counterpart in AAF, and only incurred an additional 1.23 million trainable parameters. However, increasing $r$  from 64 to 128 or from 64 to 256 resulted in  a decrease in AAF in 4/9 experiments and a minor increase in AAF in 5/9 experiments. Observing that the $r=64$ model uses the least number of parameters when compared to the $r=128$ and $r=256$ models, we elect to $r=64$ for our experiments on the CDDB dataset.

Interestingly, at the memory=500 and memory=1500 settings for the  easy and long scenarios, the LoRAX $r \geq 64$ models outperformed their full-rank implementations (DER). We hypothesize that the DER model overfit its large number of trainable parameters on the limited training data from previous tasks. We believe LoRAX did not outperform DER for the hard scenario at these memory budgets because it was the shortest sequence, with more exemplars per class preventing overfitting to limited class samples. This suggests that  by limiting training to low-rank approximations matrices LoRA, and consequently LoRAX, possibly offer regularization across tasks, mitigating forgetting on longer sequences. We also note that at the memory=100 setting, there was not a clear relationship between LoRA rank and AAF across the CDDB scenarios, possibly because of the small number of exemplar images per class.

\subsection{Testing LoRAX on ResNet}
\label{subsec:lorax_resnet}
While PEFT LoRA methods are typically applied to the attention modules of transformer architectures, we applied LoRA adapters to CNN based ResNet models as a part of our backbone model sweep for the LoRAX algorithm. We experimented with two LoRA adapter configurations 1) LoRA adapters applied to every 2D convolutional layer 2) LoRA adapters applied to the 2D convolutional layers in the last residual block (the block containing the bulk of the parameters). We applied these LoRA configurations to the ResNet34, ResNet50, and ResNet152 backbone models. Each model was tested on both LoRA configurations settings across the 9 CDDB scenario and memory combinations. On certain tasks, ResNet LoRAX performance dramatically dropped at the next task, and on others, the model was unable to achieve high initial performance. Across all scenarios and memory settings, ResNet LoRAX feature extractors either failed to retain information or to consistently extract relevant features. Therefore, we only report the LoRAX results on the ConViT architecture.

\newpage
\subsection{Additional Memory Settings}
We ran a full set of experiments across all tested CIL algorithms for memory=100 and memory=1500 exemplar budgets. Increasing exemplars improved performance across all algorithms/scenarios. DyTox performs well at the smallest memory budget of 100, where the performance gap between CNN and transformer architectures is especially apparent (up to 15\% difference). Across all scenarios, increasing the number of exemplars from 100 to 500 results in a 2-15\% increase in accuracy, especially on the long scenario, which uses the least number of exemplars per class. Increasing 500 to 1500 exemplars results in a 1-5\% increase in accuracy. LoRAX is competitive across all memory budgets/scenarios and comes out as the top performer on 2/3 CDDB scenarios at memory=500 and memory=1500.

Moreover, multiple CIL algorithms outperform the Oracle benchmark on the 1500 memory setting, possibly indicating that performance saturates around that threshold. As seen in Table \ref{table:memory_1500}, LoRAX yields a higher AA score compared to the Oracle model for all three CDDB scenarios. These results further validate our choice to approach the deepfake detection problem as a CIL task instead of showing the model the complete set of training data. 

\label{subsec:memory}
\begin{figure}[H]
\begin{center}
    \centering
    \includegraphics[scale=0.24]{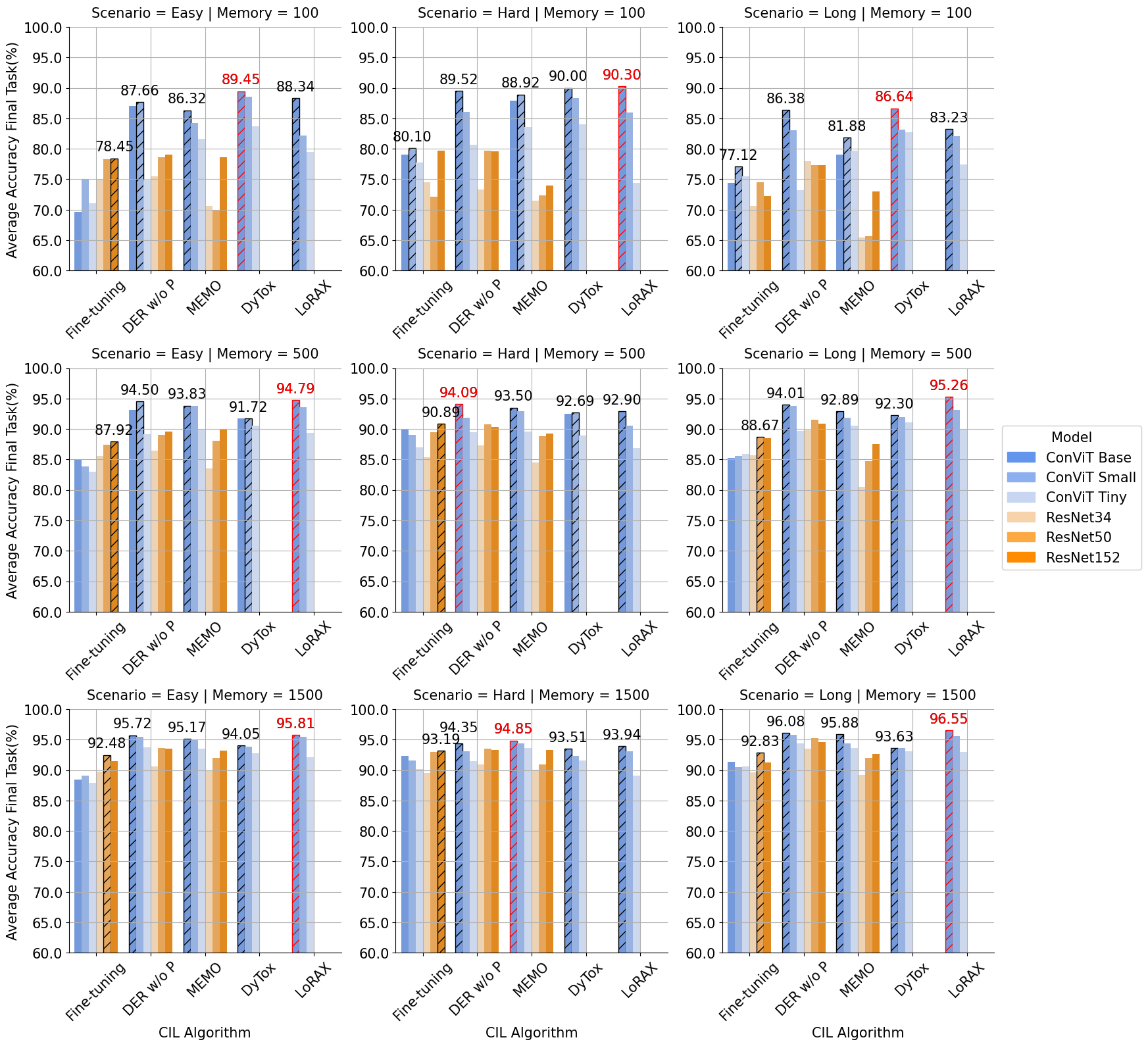} \newline
    \caption{\textbf{AAF Across CIL and Memory Settings} Black hatching: top performer within each CIL scenario; Red hatching: overall top performer for the specified scenario and memory setting.}
    \label{fig:full_bar_plots_aa}
\end{center}
\end{figure}

\newpage
\begin{figure}[ht]
    \begin{center}
        \centering
        \includegraphics[scale=0.22]{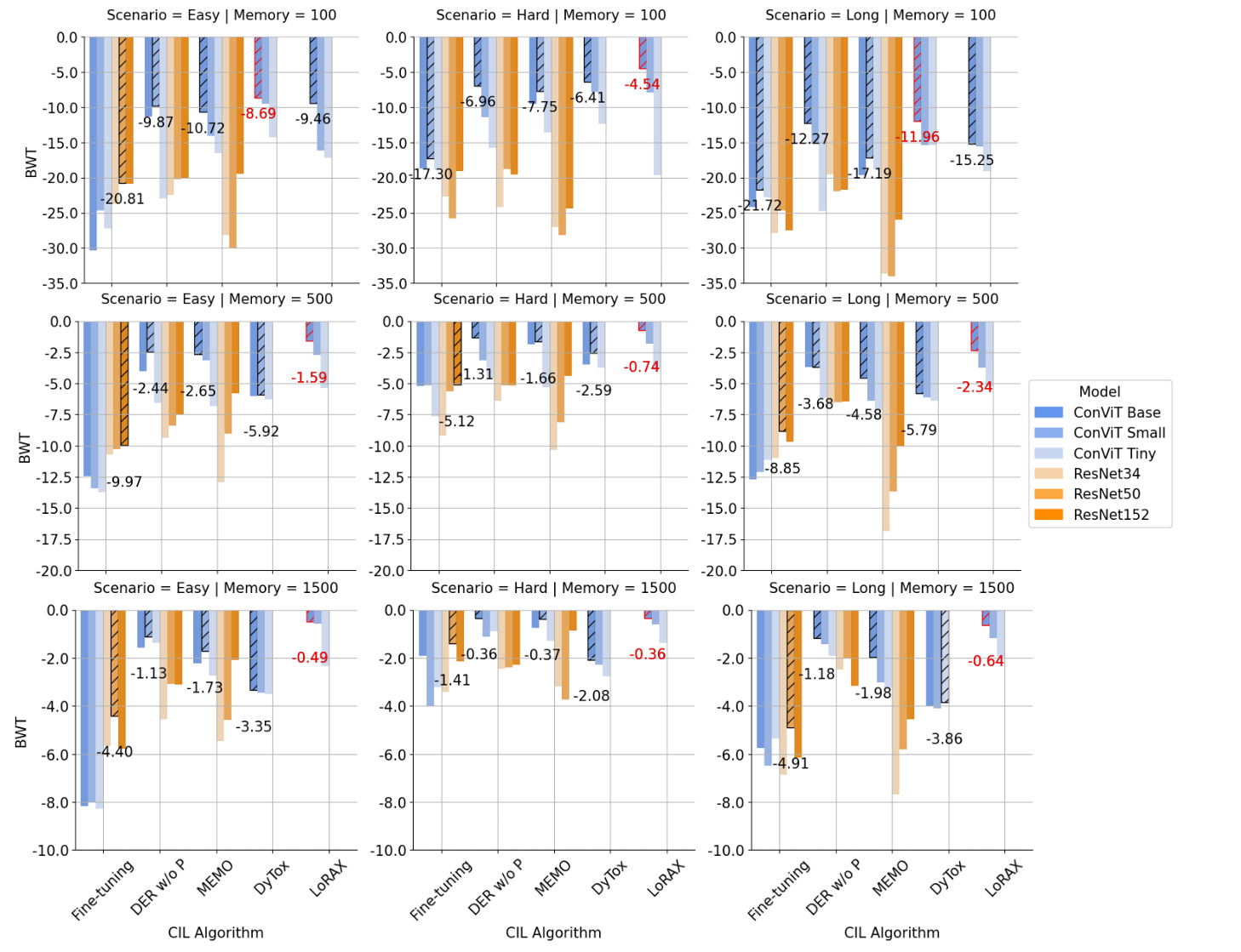} \newline
        \caption{\textbf{BWT across CIL algorithms and memory settings} Black hatching: top performer within each CIL scenario; Red hatching: overall top performer for the specified scenario and memory setting.}
        \label{fig:full_bar_plots_bwt}
    \end{center}
\end{figure}

\newpage
\begin{figure}[H]
    \begin{center}
        \centering
        \includegraphics[scale=0.22]{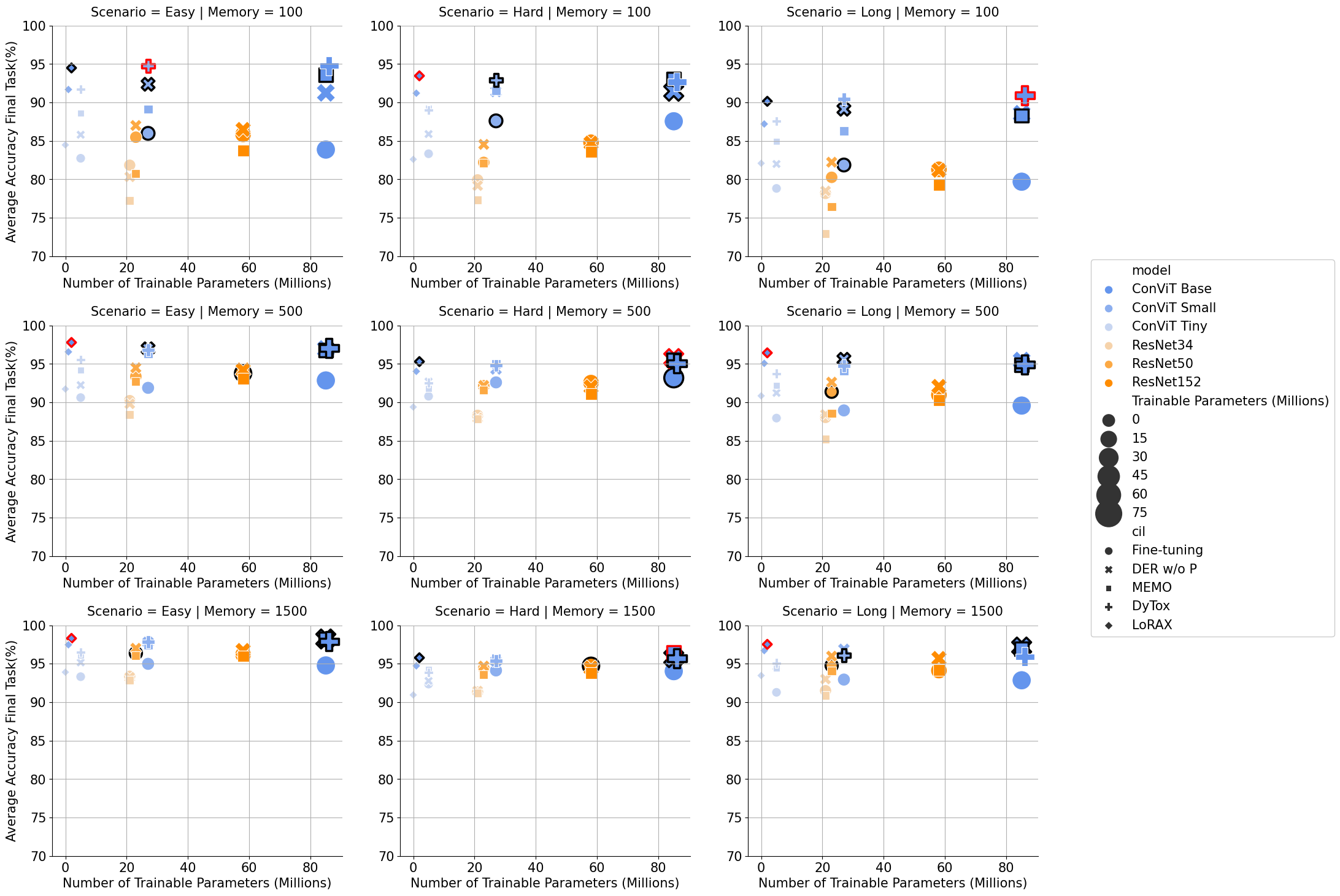} \newline
        \caption{\textbf{Average Accuracy at Final Task vs. Number of Trainable Parameters} Black outline: top performer within each CIL scenario; Red outline: overall top performer for the specified scenario and memory setting.}
        \label{fig:scatterplot_top_performers}
    \end{center}
\end{figure}

\begin{table}[H]
\begin{center}
\centering
    \begin{adjustbox}{width=0.9\textwidth}
        \begin{tabular}{c|c|cccc|| cccc || cccc}
            \hline
            {} & {} & \multicolumn{12}{c}{\textbf{100 EXEMPLARS}} \\
            \hline
            {} & {} & \multicolumn{4}{c||}{Easy (7 exemplars/class at task N)} & \multicolumn{4}{c||}{Hard (10 exemplars/class at task N)} & \multicolumn{4}{c}{Long (4 exemplars/class at task N)} \\
            \hline
             CIL & Model &    AA &    AAF &    BWT &  Params (M) &    AA &    AAF &    BWT &  Params (M) &    AA &    AAF &    BWT &  Params (M) \\
             \hline
            \multirow{6}{*}{Fine-tuning} &   ConViT Base & 83.87 & 69.60 & -30.38 &   85.80 & 87.55 & 79.05 & -18.73 &   85.80 & 79.69 & 74.39 & -24.18 &   85.80 \\
            & ConViT Small & \textbf{85.98} & 75.05 & -24.64 &   27.30 & \textbf{87.61} & \textbf{80.10} & \textbf{-17.30} &   27.30 & \textbf{81.86} & \textbf{77.12} & \textbf{-21.72} &   27.30 \\
            & ConViT Tiny & 82.74 & 71.05 & -27.23 &    5.50 & 83.33 & 77.73 & -18.90 &    5.50 & 78.82 & 75.45 & -22.76 &    5.50 \\
              &  ResNet34 & 81.84 & 74.96 & -23.68 &   21.30 & 79.94 & 74.55 & -22.68 &   21.30 & 78.16 & 70.58 & -27.90 &   21.30 \\
              &  ResNet50 & 85.48 & 78.30 & \textbf{-20.81} &   23.50 & 82.20 & 72.08 & -25.84 &   23.50 & 80.26 & 74.52 & -24.71 &   23.50 \\
            &   ResNet152 & 85.86 & \textbf{78.45} & -20.90 &   58.10 & 84.82 & 79.74 & -19.12 &   58.10 & 81.30 & 72.26 & -27.50 &   58.10 \\
            \hline
            
            \multirow{6}{*}{DER w/o P} & ConViT Base & 91.25 & 87.02 & -11.32 &   85.80 & \textbf{91.45} & \textbf{89.52} &  \textbf{-6.96} &   85.80 & 88.52 & \textbf{86.38} & \textbf{-12.27} &   85.80 \\
            & ConViT Small & \textbf{92.36} & \textbf{87.66} &  \textbf{-9.87} &   27.30 & 91.42 & 86.05 & -11.40 &   27.30 & \textbf{89.12} & 83.05 & -15.22 &   27.30 \\
             & ConViT Tiny & 85.80 & 74.86 & -22.96 &    5.50 & 85.89 & 80.66 & -15.79 &    5.50 & 82.00 & 73.26 & -24.78 &    5.50 \\
              &  ResNet34 & 80.32 & 75.52 & -22.50 &   21.30 & 79.18 & 73.27 & -24.17 &   21.30 & 78.49 & 77.99 & -19.55 &   21.30 \\
              &  ResNet50 & 87.00 & 78.60 & -20.18 &   23.50 & 84.54 & 79.73 & -18.84 &   23.50 & 82.23 & 77.33 & -21.88 &   23.50 \\
              & ResNet152 & 86.48 & 79.03 & -20.00 &   58.10 & 84.71 & 79.63 & -19.52 &   58.10 & 81.18 & 77.30 & -21.75 &   58.10 \\
            \hline
              
             \multirow{6}{*}{MEMO} & ConViT Base & \textbf{93.57} & \textbf{86.32} & \textbf{-10.72} &   85.80 & \textbf{93.07} & 87.89 &  -9.54 &   85.80 & \textbf{88.29} & 78.99 & -19.64 &   85.80 \\
            & ConViT Small & 89.12 & 84.25 & -14.11 &   27.30 & 91.53 & \textbf{88.92} &  \textbf{-7.75} &   27.30 & 86.27 & \textbf{81.88} & \textbf{-17.19} &   27.30 \\
             & ConViT Tiny & 88.58 & 81.59 & -16.53 &    5.50 & 89.21 & 83.61 & -13.64 &    5.50 & 84.95 & 79.68 & -18.76 &    5.50 \\
              &  ResNet34 & 77.22 & 70.62 & -28.18 &   21.30 & 77.34 & 71.53 & -27.05 &   21.30 & 72.91 & 65.43 & -33.67 &   21.30 \\
              &  ResNet50 & 80.78 & 69.98 & -30.05 &   23.50 & 82.10 & 72.36 & -28.17 &   23.50 & 76.42 & 65.65 & -34.06 &   23.50 \\
              & ResNet152 & 83.74 & 78.63 & -19.48 &   58.10 & 83.58 & 73.95 & -24.41 &   58.10 & 79.29 & 73.05 & -26.02 &   58.10 \\
            \hline
              
             \multirow{3}{*}{DyTox} & ConViT Base & \textbf{\textcolor{red}{\hl{94.78}}} & \textbf{\textcolor{red}{\hl{89.45}}} &  \textbf{\textcolor{red}{\hl{-8.69}}} &   64.50 & 92.68 & \textbf{90.00} &  \textbf{-6.41} &   64.50 & \textbf{\textcolor{red}{\hl{90.92}}} & \textbf{\textcolor{red}{\hl{86.64}}} & \textbf{\textcolor{red}{\hl{-11.96}}} &   64.50 \\
            & ConViT Small & \textbf{94.78} & 88.50 &  -9.53 &   20.90 & \textbf{92.89} & 88.35 &  -7.89 &   20.90 & 90.45 & 83.17 & -15.38 &   20.90 \\
            & ConViT Tiny & 91.74 & 83.70 & -14.28 &    4.14 & 89.00 & 84.05 & -12.37 &    4.14 & 87.59 & 82.74 & -15.37 &    4.14 \\
            \hline
            
             \multirow{3}{*}{LoRAX} & ConViT Base & \textbf{94.49} & \textbf{88.34} &  \textbf{-9.46} &    \textbf{\textcolor{red}{2.50}} & \textbf{\textcolor{red}{\hl{93.46}}} & \textbf{\textcolor{red}{\hl{90.30}}} &  \textbf{\textcolor{red}{\hl{-4.54}}} &    \textbf{\textcolor{red}{2.50}} & \textbf{90.15} & \textbf{83.23} & \textbf{-15.25} &   \textbf{\textcolor{red}{2.50}} \\
            & ConViT Small & 91.68 & 82.21 & -16.19 &   \textbf{\textcolor{red}{1.40}} & 91.20 & 86.01 &  -7.91 &   \textbf{\textcolor{red}{1.40}} & 87.20 & 82.11 & -15.57 &   \textbf{\textcolor{red}{1.40}} \\
            & ConViT Tiny & 84.48 & 79.48 & -17.18 &   \textbf{\textcolor{red}{0.60}} & 82.60 & 74.36 & -19.65 &   \textbf{\textcolor{red}{0.60}} & 82.09 & 77.47 & -19.09 &   \textbf{\textcolor{red}{0.60}} \\
            \hline

            {} & {} & \multicolumn{12}{c}{\textbf{ALL TRAINING DATA }}\\
        \hline
        \multirow{6}{*}{Oracle}& 
         ConViT Base & 96.52 & 93.99 & N/A &   85.80 & 93.92 & 91.47 & N/A &   85.80 & 95.71 & 94.28 & N/A &   85.80 \\
        & ConViT Small & 96.71 & 94.38 & N/A &   27.30 & 94.19 & 91.68 &  N/A &   27.30 & 95.99 & 94.36 & N/A &   27.30 \\
         & ConViT Tiny & 96.30 & 93.81 & N/A &    5.50 & 93.62 & 91.64 & N/A &    5.50 & 95.60 & 94.27 & N/A &    5.50 \\
          &  ResNet34 & 95.87 & 94.07 & N/A &   21.30 & 93.71 & 91.82 &  N/A &   21.30 & 95.14 & 94.11 & N/A &   21.30 \\
          &  ResNet50 & \textbf{\textcolor{blue}{98.11}} & \textbf{\textcolor{blue}{95.83}} & N/A &   23.50 & \textbf{\textcolor{blue}{95.69}} & 93.87 & N/A &   23.50 & 96.74 & 94.66 & N/A &   23.50 \\
          &  ResNet152 & 97.70 & 95.44 & N/A &   58.10 & 95.65 & \textbf{\textcolor{blue}{95.24}} &  N/A &   58.10 & \textbf{\textcolor{blue}{96.77}} & \textbf{\textcolor{blue}{95.34}} & N/A &   58.10 \\
        \hline
        \end{tabular}
    \end{adjustbox}
    \end{center}
\vspace{0.5cm}
\caption{Results on our LoRAX method benchmarked against other CIL algorithms with 100 exemplar images across all classes.}
\label{table:memory_100}
\end{table}
\begin{table}[H]
\begin{center}
\centering
    \begin{adjustbox}{width=0.9\textwidth}
        \begin{tabular}{c|c|cccc|| cccc || cccc}
            \hline
            {} & {} & \multicolumn{12}{c}{\textbf{1500 EXEMPLARS}} \\
            \hline
            {} & {} & \multicolumn{4}{c||}{Easy (107 exemplars/class at task N)} & \multicolumn{4}{c||}{Hard (150 exemplars/class at task N)} & \multicolumn{4}{c}{Long (62 exemplars/class at task N)} \\
            \hline
             CIL & Model &    AA &    AAF &    BWT &  Params (M) &    AA &    AAF &    BWT &  Params (M) &    AA &    AAF &    BWT &  Params (M) \\
             \hline
            \multirow{6}{*}{Fine-tuning} &  ConViT Base & 94.78 & 88.49 & -8.17 &   85.80 & 94.03 & 92.38 & -1.91 &   85.80 & 92.85 & 91.35 & -5.75 &   85.80 \\
            & ConViT Small & 94.99 & 89.10 & -8.00 &   27.30 & 94.13 & 91.53 & -4.01 &   27.30 & 92.94 & 90.53 & -6.50 &   27.30 \\
             & ConViT Tiny & 93.32 & 87.93 & -8.28 &    5.50 & 92.36 & 90.12 & -3.22 &    5.50 & 91.28 & 90.58 & -5.36 &    5.50 \\
               & ResNet34 & 93.38 & 89.73 & -5.66 &   21.30 & 91.33 & 89.57 & -3.43 &   21.30 & 91.50 & 89.68 & -6.86 &   21.30 \\
               & ResNet50 & \textbf{96.37} & \textbf{92.48} & \textbf{-4.40} &   23.50 & 94.41 & 92.95 & \textbf{-1.41} &   23.50 & \textbf{94.78} & \textbf{92.83} & \textbf{-4.91} &   23.50 \\
              & ResNet152 & 96.19 & 91.47 & -5.80 &   58.10 & \textbf{94.73} & \textbf{93.19} & -2.16 &   58.10 & 94.11 & 91.23 & -6.14 &   58.10 \\
            \hline
               
             \multirow{6}{*}{DER w/o P} & ConViT Base & \textbf{98.22} & \textbf{95.72} & -1.58 &   85.80 & \textbf{95.79} & \textbf{94.35} & \textbf{\textcolor{red}{\hl{-0.36}}} &   85.80 & \textbf{97.17} & \textbf{96.08} & \textbf{-1.18} &   85.80 \\
            & ConViT Small & 97.78 & 95.52 & \textbf{-1.13} &   27.30 & 95.35 & 93.12 & -1.12 &   27.30 & 96.87 & 95.77 & -1.43 &   27.30 \\
            & ConViT Tiny & 95.17 & 93.73 & -1.36 &    5.50 & 92.77 & 91.51 & -0.90 &    5.50 & 94.62 & 94.39 & -1.91 &    5.50 \\
             &   ResNet34 & 93.25 & 90.60 & -4.56 &   21.30 & 91.45 & 90.88 & -2.46 &   21.30 & 93.01 & 93.54 & -2.50 &   21.30 \\
              &  ResNet50 & 97.01 & 93.65 & -3.08 &   23.50 & 94.69 & 93.49 & -2.41 &   23.50 & 95.99 & 95.22 & -2.02 &   23.50 \\
              & ResNet152 & 96.72 & 93.48 & -3.10 &   58.10 & 94.44 & 93.34 & -2.29 &   58.10 & 95.68 & 94.58 & -3.17 &   58.10 \\
                \hline
               
             \multirow{6}{*}{MEMO} & ConViT Base & \textbf{98.22} & \textbf{95.17} & -2.23 &   85.80 & \textbf{\textcolor{red}{\hl{96.46}}} & \textbf{\textcolor{red}{\hl{94.85}}} & -0.76 &   85.80 & \textbf{96.94} & \textbf{95.88} & \textbf{-1.98} &   85.80 \\
            & ConViT Small & 97.47 & 94.97 & \textbf{-1.73} &   27.30 & 95.69 & 94.40 & \textbf{-0.37} &   27.30 & 96.19 & 94.40 & -3.03 &   27.30 \\
             & ConViT Tiny & 96.10 & 93.54 & -2.76 &    5.50 & 94.27 & 93.58 & -1.30 &    5.50 & 94.42 & 93.59 & -3.26 &    5.50 \\
               & ResNet34 & 92.88 & 89.97 & -5.47 &   21.30 & 91.19 & 90.07 & -3.20 &   21.30 & 90.85 & 89.17 & -7.70 &   21.30 \\
               & ResNet50 & 96.05 & 91.97 & -4.59 &   23.50 & 93.59 & 90.89 & -3.72 &   23.50 & 94.07 & 91.99 & -5.80 &   23.50 \\
              & ResNet152 & 95.98 & 93.19 & -2.10 &   58.10 & 93.80 & 93.30 & -0.86 &   58.10 & 94.17 & 92.65 & -4.56 &   58.10 \\
                \hline
               
             \multirow{3}{*}{DyTox} & ConViT Base & \textbf{97.90} & \textbf{94.05} & \textbf{-3.35} &   64.50 & \textbf{95.70} & \textbf{93.51} & \textbf{-2.08} &   64.50 & 95.90 & \textbf{93.63} & \textbf{-4.01} &   64.50 \\
            & ConViT Small & 97.83 & 93.88 & -3.46 &   20.90 & 95.40 & 92.29 & -2.28 &   20.90 & \textbf{96.13} & 93.59 & -4.10 &   20.90 \\
             & ConViT Tiny & 96.50 & 92.71 & -3.52 &    4.14 & 93.88 & 91.55 & -2.76 &    4.14 & 95.09 & 93.14 & -3.86 &    4.14 \\
            \hline
             
             \multirow{3}{*}{LoRAX} & ConViT Base & \textbf{\textcolor{red}{\hl{98.30}}} & \textbf{\textcolor{red}{\hl{95.81}}} & \textbf{\textcolor{red}{\hl{-0.49}}} &    2.50 & \textbf{95.78} & \textbf{93.94} & \textbf{\textcolor{red}{\hl{-0.36}}} &    2.50 & \textbf{\textcolor{red}{\hl{97.52}}} & \textbf{\textcolor{red}{\hl{96.55}}} & \textbf{\textcolor{red}{\hl{-0.64}}} &    \textbf{\textcolor{red}{2.50}} \\
             & ConViT Small & 97.47 & 95.42 & -0.57 &    1.40 & 94.72 & 93.04 & -0.60 &    1.40 & 96.71 & 95.61 & -1.16 &    \textbf{\textcolor{red}{1.40}} \\
             & ConViT Tiny & 93.90 & 92.11 & -2.33 &    0.60 & 90.94 & 89.07 & -1.39 &    0.60 & 93.47 & 92.93 & -1.98 &    \textbf{\textcolor{red}{0.60}} \\
            \hline
        {} & {} & \multicolumn{12}{c}{\textbf{ALL TRAINING DATA }}\\
        \hline
        \multirow{6}{*}{Oracle}& 
         ConViT Base & 96.52 & 93.99 & N/A &   85.80 & 93.92 & 91.47 & N/A &   85.80 & 95.71 & 94.28 & N/A &   85.80 \\
        & ConViT Small & 96.71 & 94.38 & N/A &   27.30 & 94.19 & 91.68 &  N/A &   27.30 & 95.99 & 94.36 & N/A &   27.30 \\
         & ConViT Tiny & 96.30 & 93.81 & N/A &    5.50 & 93.62 & 91.64 & N/A &    5.50 & 95.60 & 94.27 & N/A &    5.50 \\
          &  ResNet34 & 95.87 & 94.07 & N/A &   21.30 & 93.71 & 91.82 &  N/A &   21.30 & 95.14 & 94.11 & N/A &   21.30 \\
          &  ResNet50 & \textbf{\textcolor{blue}{98.11}} & \textbf{\textcolor{blue}{95.83}} & N/A &   23.50 & \textbf{\textcolor{blue}{95.69}} & 93.87 & N/A &   23.50 & 96.74 & 94.66 & N/A &   23.50 \\
          &  ResNet152 & 97.70 & 95.44 & N/A &   58.10 & 95.65 & \textbf{\textcolor{blue}{95.24}} &  N/A &   58.10 & \textbf{\textcolor{blue}{96.77}} & \textbf{\textcolor{blue}{95.34}} & N/A &   58.10 \\
        \hline
        \end{tabular}
    \end{adjustbox}
    \end{center}
    \vspace{0.5cm}
    \caption{Results on our LoRAX method benchmarked against other CIL algorithms with 1500 exemplar images across all classes.}
    \label{table:memory_1500}
\end{table}

\subsection{Total Parameters and Backbone Memory Storage}
\begin{table}[H]
\begin{center}
\centering
    \begin{adjustbox}{width=0.9\textwidth}
        \begin{tabular}{c|c|ccc|| ccc || ccc}
            \hline
            {} & {} & \multicolumn{9}{c}{\textbf{PARAMETERS AND MEMORY STORAGE}} \\
            \hline
            {} & {} & \multicolumn{3}{c||}{Easy} & \multicolumn{3}{c||}{Hard} & \multicolumn{3}{c}{Long} \\
            \hline
            CIL & Model &  Trainable (M) &  Total (M) &  Model Size (MB) &  Trainable (M) &  Total (M) &  Model Size (MB) &  Trainable (M) &  Total (M) &  Model Size (MB) \\
            \hline
            \multirow{6}{*}{Fine-tuning} & ConViT Base &        85.8 &          85.8 &            327 &        85.8 &          85.8 &            327 &        85.8 &          85.8 &            327 \\
            &ConViT Small &        27.3 &          27.4 &            104 &        27.3 &          27.3 &            104 &        27.3 &          27.4 &            104 \\
             &ConViT Tiny &         5.5 &           5.5 &             21 &         5.5 &           5.5 &             21 &         5.5 &           5.5 &             21 \\
             &   ResNet34 &        21.3 &          21.3 &             81 &        21.3 &          21.3 &             81 &        21.3 &          21.3 &             81 \\
              &  ResNet50 &        23.5 &          23.6 &             89 &        23.5 &          23.6 &             89 &        23.5 &          23.6 &             90 \\
              & ResNet152 &        58.1 &          58.3 &            222 &        58.1 &          58.3 &            222 &        58.1 &          58.3 &            222 \\
            \hline
              
             \multirow{6}{*}{DER w/o P} & ConViT Base &        85.8 &         600.5 &           2290 &        85.8 &         428.9 &           1636 &        85.8 &        1029.5 &           3927 \\
            & ConViT Small &        27.3 &         191.5 &            730 &        27.3 &         136.7 &            521 &        27.3 &         328.3 &           1252 \\
             & ConViT Tiny &         5.5 &          38.6 &            147 &         5.5 &          27.6 &            105 &         5.5 &          66.3 &            252 \\
               & ResNet34 &        21.3 &         149.2 &            569 &        21.3 &         106.5 &            406 &        21.3 &         255.8 &            975 \\
               & ResNet50 &        23.5 &         165.1 &            629 &        23.5 &         117.9 &            449 &        23.5 &         283.3 &           1080 \\
              & ResNet152 &        58.1 &         408.3 &           1557 &        58.1 &         291.6 &           1112 &        58.1 &         700.1 &           2670 \\
            \hline
            
            \multirow{6}{*}{MEMO} &   ConViT Base &        85.8 &         128.4 &            489 &        85.8 &         114.2 &            435 &        85.8 &         164.0 &            625 \\
            & ConViT Small &        27.3 &          40.9 &            155 &        27.3 &          36.3 &            138 &        27.3 &          52.2 &            198 \\
            & ConViT Tiny &         5.5 &           8.2 &             31 &         5.5 &           7.3 &             27 &         5.5 &          10.5 &             39 \\
            &    ResNet34 &        21.3 &         100.1 &            381 &        21.3 &          73.8 &            281 &        21.3 &         165.8 &            632 \\
            &    ResNet50 &        23.5 &         113.7 &            433 &        23.5 &          83.6 &            319 &        23.5 &         189.0 &            721 \\
            &   ResNet152 &        58.1 &         148.4 &            566 &        58.1 &         118.4 &            451 &        58.1 &         223.8 &            853 \\
            \hline
           
           \multirow{3}{*}{DyTox} &   ConViT Base &        64.5 &          64.5 &            246 &        64.5 &          64.5 &            246 &        64.6 &          64.6 &            246 \\
           &   ConViT Small &        20.5 &          20.5 &            78 &        20.5 &          20.5 &            78 &        20.5 &          20.6 &            78 \\
           &   ConViT Tiny &        4.1 &          4.1 &            16 &        4.1 &          4.1 &            16&        4.1 &          4.1 &            16\\
            \hline
             
             \multirow{3}{*}{LoRAX} &  ConViT Base &         2.5 &         103.1 &            393 &         2.5 &          98.1 &            374 &         2.5 &         115.5 &            440 \\
            & ConViT Small &         1.4 &          37.1 &            141 &         1.4 &          34.3 &            130 &         2.5 &         115.5 &            440 \\
            & ConViT Tiny &         0.6 &           9.8 &             37 &         0.6 &           8.6 &             32 &         0.6 &          12.9 &             49 \\
            \hline
        \end{tabular}
    \end{adjustbox}
    \end{center}
    \vspace{0.5cm}
    \caption{Trainable parameters, total parameters in millions (M), and model size in MB of each CIL and CDDB scenario. The number of parameters are equal across memory settings.}
    \label{table:params_memory}
\end{table}

Compared to the DER w/o P model, which uses feature concatenation with full-rank weight matrices, LoRAX utilizes 2.9\% of the trainable parameters for ConViT Base, 5.1\% for ConViT Small, and 10.9\% for ConViT Tiny. The total number of parameters was calculated using the stored model checkpoint at the last task for each of the easy, hard, and long tasks and was converted into MB based on the size of a float (4 bytes).

\subsection{LoRAX Parameter Reductions}

By applying our recommended LoRA adapter configuration, we effectively approximated high-rank ConViT weight updates using low-rank matrices, resulting in a reduction of trainable parameters by $\sim35$ times. In particular, the number of trainable parameters for the ConViT Base model, the largest and top-performing model, decreased from $\sim86$ million to $\sim2.5$ million trainable parameters after applying the suggested LoRAX configuration. To quantify the number of parameters in terms of exemplar images, consider that the memory it takes to store a single $224\times224$ pixel exemplar image corresponds to 37,362 parameters, where each parameter is a 32-bit/4-byte floating point value. Hence, adding an adapter per task for the LoRAX ConViT Base model with rank $r = 64$ corresponds to storing just 65  exemplar images, a useful trade-off for resource-constrained environments.


\end{document}